\documentclass[preprint,12pt]{elsarticle}

\usepackage{amssymb}
\usepackage{url}
\usepackage{hyperref}
\usepackage{amsmath}
\usepackage{algorithm}
\usepackage{algpseudocode}
\usepackage{multirow}
\usepackage{booktabs}
\usepackage{lineno}

\journal{ISPRS Journal of Photogrammetry and Remote Sensing}

\begin{document}
\begin{frontmatter}
\title{TSE-Net: Semi-supervised Monocular Height Estimation from Single Remote Sensing Images}

\author[label1,label2]{Sining Chen}
\author[label1,label2]{Xiao Xiang Zhu}
\affiliation[label1]{organization={Chair of Data Science in Earth Observation, Technical University of Munich (TUM)},
            city={Munich},
            postcode={80333},
            country={Germany}}

\affiliation[label2]{organization={Munich Center for Machine Learning (MCML)},
            city={Munich},
            postcode={80538},
            country={Germany}}

\begin{abstract}
Monocular height estimation plays a critical role in 3D perception for remote sensing, offering a cost-effective alternative to multi-view or LiDAR-based methods. While deep learning has significantly advanced the capabilities of monocular height estimation, these methods remain fundamentally limited by the availability of labeled data, which are expensive and labor-intensive to obtain at scale. The scarcity of high-quality annotations hinders the generalization and performance of existing models.
To overcome this limitation, we propose leveraging large volumes of unlabeled data through a semi-supervised learning framework, enabling the model to extract informative cues from unlabeled samples and improve its predictive performance.
In this work, we introduce TSE-Net, a self-training pipeline for semi-supervised monocular height estimation. The pipeline integrates teacher, student, and exam networks. The student network is trained on unlabeled data using pseudo-labels generated by the teacher network, while the exam network functions as a temporal ensemble of the student network to stabilize performance. The teacher network is formulated as a joint regression and classification model: the regression branch predicts height values that serve as pseudo-labels, and the classification branch predicts height value classes along with class probabilities, which are used to filter pseudo-labels. Height value classes are defined using a hierarchical bi-cut strategy to address the inherent long-tailed distribution of heights, and the predicted class probabilities are calibrated with a Plackett-Luce model to reflect the expected accuracy of pseudo-labels. We evaluate the proposed pipeline on three datasets spanning different resolutions and imaging modalities: the ISPRS Vaihingen dataset (IR-R-G aerial images, 0.09 m resolution), the SynRS3D dataset (synthetic images, 0.6--1 m resolution), and the GBH dataset (satellite RGB images, 3 m resolution). Experimental results demonstrate that our approach consistently outperforms baseline methods, and ablation studies confirm the contribution of each design component. Codes are available at \url{https://github.com/zhu-xlab/tse-net}. 
\end{abstract}



\end{frontmatter}
\section{Introduction}
Monocular height estimation from single remote sensing images produces height maps delivered as normalized digital surface models (nDSMs). These maps are crucial for a wide range of downstream applications, including 3D building modeling in urban areas \cite{cao2021,cao2024} and canopy height models in forests \cite{lang2023,fayad2024}. As buildings are among the most prominent man-made structures, their 3D representations enhance our understanding of anthropogenic processes such as urbanization \cite{taubenbock2012}, population dynamics \cite{biljecki2016}, and energy exchange \cite{li2023d}. Forests play a vital role in the global carbon cycle, and therefore, accurate canopy height models are indispensable for studying and mitigating climate change \cite{bonan2008}.

Height maps can also be derived from alternative techniques, such as direct 3D observations with light detection and ranging (LiDAR) \cite{priestnall2000,chen2022a}, synthetic aperture radar (SAR) \cite{zhu2014,zhu2014a}, or stereo pairs of optical images \cite{dangelo2008}. Nevertheless, single-image-based approaches offer distinct advantages, including lower costs, reduced occlusions, and improved data availability. The abundance of single-view imagery not only facilitates large-scale applications with timely updates but also enables deep learning models to leverage a vast amount of data for improved generalization.

Deep learning provides a data-driven solution to monocular height estimation, eliminating the need for handcrafted features and offering greater flexibility and representational capacity. However, the performance of such models heavily depends on the quantity and quality of labeled data. While numerous studies have explored deep-learning-based monocular height estimation \cite{srivastava2017,mou2018,ghamisi2018,amirkolaee2019a,carvalho2020,elhousni2021,xing2021,li2020b,chen2023e}, most rely on fully supervised learning, assuming the availability of ground-truth labels for every sample. In practice, labeled data are scarce \cite{thebenchmark,xiong2024b}, as height information is typically acquired from costly 3D sensors such as LiDAR or stereo systems.

Semi-supervised learning (SSL) offers a promising solution by exploiting unlabeled data to guide network training. Among SSL techniques, consistency-based \cite{ouali2020,liu2022a,wang2023g,wang2024a,mai2024} and self-training-based \cite{sohn2020a,zhang2022,chen2023i,sun2024,yang2023d} approaches have achieved state-of-the-art results and also proven effective in remote sensing applications \cite{wang2020d,zhang2023d,li2022c,zhang2023c,huang2023,huang2024,huang2025}. However, these methods cannot be directly applied to semi-supervised monocular height estimation. First, most of them are designed for classification tasks, either image-level classification or pixel-level segmentation, where the discrete label space is inherently robust to false regularization signals from the teacher network. In contrast, for height regression, pseudo-label filtering becomes crucial to prevent error propagation. Second, height values exhibit an extremely long-tailed distribution \cite{chen2023g}, and teacher networks tend to neglect rare high-value pixels, further exacerbating underestimation in the student network.

To overcome these limitations, we propose a novel semi-supervised strategy specifically designed for monocular height estimation. Inspired by previous work that reformulates height estimation as a classification or hybrid classification-regression task \cite{li2020b,chen2023e}, we introduce a pseudo-label filtering mechanism based on an uncertainty proxy derived from the class probabilities of height values. To ensure that this proxy reliably reflects pseudo-label accuracy, we employ a Plackett-Luce model \cite{plackett1975,luce2005} to align the ranking of class probabilities with pseudo-label errors. Furthermore, to address the long-tailed distribution of height values, we design a hierarchical bi-cut strategy that balances class samples across the entire value range and emphasizes high-value pixels during training.

In summary, this work presents the first study of semi-supervised learning for monocular height estimation. Our main contributions are as follows:
\begin{itemize}
\item We propose the first semi-supervised framework for monocular height estimation, introducing a self-training pipeline named TSE-Net (Teacher-Student-Exam). Unlike conventional self-training approaches, TSE-Net incorporates an additional exam sub-network that temporally aggregates the student network for improved performance stability.
\item We design a joint classification-regression network as the teacher network, which generates both pseudo-labels and confidence metrics. The confidence metrics are aligned with pseudo-label errors through a Plackett-Luce model.
\item We mitigate the long-tailed height distribution via a hierarchical bi-cut strategy that balances class samples and emphasizes high-value pixels, thereby reducing systematic underestimation.
\item We conduct extensive experiments and ablation studies, demonstrating the effectiveness of the proposed pipeline and validating the contribution of each design component.
\end{itemize}

The remainder of the article is organized as follows. Section \ref{chap:related_works} reviews related studies. The proposed method is described in detail in Section \ref{chap:methodology}, followed by Section \ref{chap:experiments}, which describes the experiments, and Section \ref{chap:results}, which presents the experimental results. Ablation studies and discussions are presented in Section \ref{chap:discussions}. Conclusions are drawn, and further research directions are described in Section \ref{chap:conclusions}.
\section{Related Works}\label{chap:related_works}
\subsection{Monocular Height Estimation}
Monocular height estimation takes a single image as input and outputs a corresponding height map. In essence, it is a dense prediction problem that can be approached through three different paradigms: regression, classification, and classification-regression.

\subsubsection{Regression}
Treating monocular height estimation as a regression task is the most straightforward approach, where the output space is continuous. Most methods in this category adopt encoder-decoder fully convolutional networks (FCN). As a dense regression problem, many off-the-shelf FCNs originally designed for semantic segmentation can be adapted by removing the output activation, such as FCN \cite{shelhamer2017}, SegNet \cite{badrinarayanan2017}, U-Net \cite{ronneberger2015}, and Eff-UNet \cite{baheti2020}. 

Several networks have been developed specifically for monocular height estimation \cite{srivastava2017,mou2018,ghamisi2018,amirkolaee2019a,carvalho2020,elhousni2021,xing2021}. For instance, one of the earliest neural networks for this task was proposed by Mou \& Zhu \cite{mou2018}, featuring a standard encoder-decoder FCN architecture with skip connections for low-level feature fusion. Amirkolaee et al. \cite{amirkolaee2019a} improved performance through multi-level feature fusion, and reduced computational cost with an efficient up-sampling block. Xing \textit{et al}. \cite{xing2021} introduced a gated feature aggregation module (GFAM) for feature fusion and a progressive refinement module (PRM) for result refinement. Additionally, Srivastava et al. \cite{srivastava2017} explored multi-task learning by jointly predicting height and semantic segmentation, demonstrating the benefit of incorporating semantic cues.

With the recent emergence of foundation models in computer vision, several large-scale approaches have been developed for monocular depth estimation (MDE), a closely related task. For example, DepthAnything \cite{yang2024c,yang2024d} introduced the first MDE foundation model, built upon DINO encoders \cite{oquab2024} and trained on massive datasets, showing strong generalizability across diverse scenes. More recently, Marigold \cite{ke2025} leveraged diffusion models, enabling efficient adaptation for single-image depth inference.

\subsubsection{Classification}
Inspired by advances in MDE, researchers have explored reframing monocular height estimation as a classification problem. Fu et al. \cite{fu2018a} first proposed discretizing the continuous output space into a set of ordinal classes, where each class corresponds to a height interval. The model predicts the class for each pixel, and ordinal regression is introduced to preserve the inherent ordering among classes, leading to improved performance. Li et al. \cite{li2020b} extended this paradigm to remote sensing, demonstrating its feasibility for monocular height estimation.

Although this classification-based formulation achieved state-of-the-art performance at that time, it introduced quantization artifacts because the final height values are typically approximated by the midpoints of the predicted class intervals.
\subsubsection{Classification-Regression}
To mitigate the discretization artifacts of pure classification methods, classification-regression approaches have been developed. AdaBins \cite{bhat2021} and its successors \cite{bhat2022,li2024a} combined the strengths of both paradigms by estimating height values as the expected value of class probability distributions. Moreover, they introduced adaptive bins, where class boundaries dynamically adjust to the image-specific value distribution. Inspired by these advances, Chen et al. \cite{chen2023e} proposed HTC-DC Net, which further addressed the long-tailed distribution of height values and constrains the pixel-wise distribution, achieving state-of-the-art results in monocular height estimation.

Despite these substantial advances, existing methods generally assume the availability of sufficient labeled data for supervised training. In practice, acquiring accurate height labels is expensive and labor-intensive, underscoring the need for methods that perform well with limited annotations.
\subsection{Semi-supervised Learning}
Semi-supervised learning (SSL) addresses scenarios where labeled data samples are restrictively available but unlabeled data are abundant. State-of-the-art approaches are broadly categorized into consistency-based and self-training-based approaches. 

Consistency regularization enforces that the model consistently produces predictions on the same inputs under different perturbations \cite{ouali2020,liu2022a,wang2023g,wang2024a,mai2024}. For example, Ouali et al. \cite{ouali2020} proposed cross-consistency training, which regularizes intermediate representations under multiple perturbations. Wang et al. \cite{wang2024a} conducted a feature-level consistency learning framework called Density-Descending Feature Perturbation (DDFP), which aims at effective regularization of the low-density decision boundary. Mai et al. \cite{mai2024} proposed RankMatch, which incorporates inter-pixel correlation under perturbations. In remote sensing, Wang et al. \cite{wang2020d} first utilized consistency-based semi-supervised learning for semantic segmentation. Zhang et al. \cite{zhang2023d} proposed a transformation consistency regularization framework that considers various spatial transformations or perturbations. Li et al. \cite{li2022c} focused on feature-level perturbations for semi-supervised building footprint generation.

On the other hand, self-training expands the effective training set by including pseudo-labels for the unlabeled data samples \cite{sohn2020a,zhang2022,chen2023i,sun2024,yang2023d}. FixMatch \cite{sohn2020a} is one of the most influential self-training semi-supervised learning frameworks, where weak and strong augmentations are applied to create two distinct views of the same image, and pseudo-labels from the weak view supervise the strong one. In this context, pseudo-label filtering is crucial. FixMatch conducts filtering via a fixed threshold, whereas later works, such as FlexMatch \cite{zhang2022} and SoftMatch \cite{chen2023i}, introduced adaptive or dynamic thresholds for different classes. In remote sensing, Zhang et al. \cite{zhang2023c} employed a joint self-training pipeline for semi-supervised change detection. Huang et al. extended self-training for semantic segmentation with adaptive matching \cite{huang2023}, decoupling and weighting strategies \cite{huang2024}, and debiasing mechanisms \cite{huang2025}.

Despite these advantages, the SSL approaches are developed for classification or segmentation tasks with discrete outputs. In contrast, regression problems such as monocular height estimation remain underexplored. This gap motivates the development of dedicated SSL frameworks tailored for continuous-value prediction tasks.
\section{Methodology}\label{chap:methodology}
\begin{figure*}[!htbp]
    \centering
    \includegraphics[width=\linewidth]{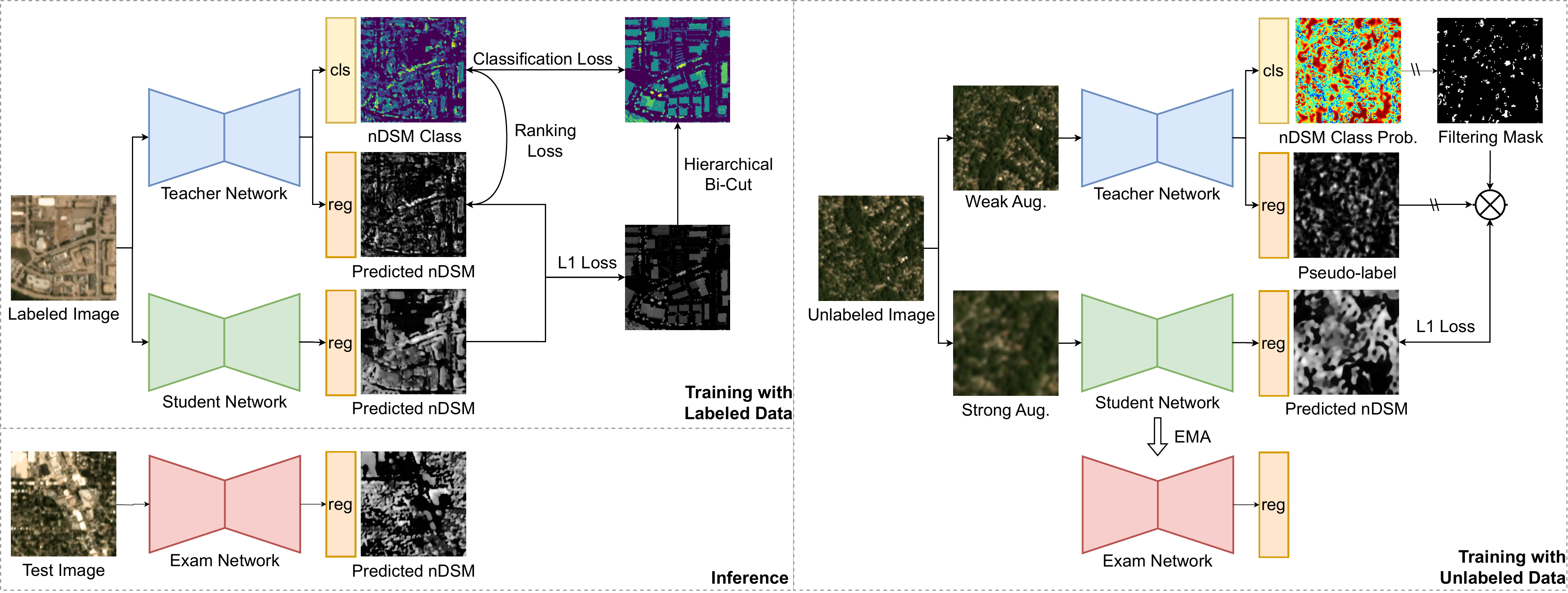}
    \caption{TSE-Net Pipeline. The framework consists of three networks: the teacher (T), student (S), and exam (E) networks. The teacher network is a classification-regression multi-task network, where the class probabilities are calibrated to align with the regression errors via a Plackett-Luce model. The student and exam networks share the same regression architecture. On the labeled dataset, the teacher and student networks are jointly trained using the ground truth labels. On the unlabeled dataset, two views---a weak and a strong augmentation---are fed into the teacher and student networks, respectively. The teacher produces pseudo-labels that are filtered by class probabilities to supervise the student network. The exam network is updated as the exponential moving average (EMA) of the student network. During inference, only the exam network is used.}
    \label{fig:architecture}
\end{figure*}
As illustrated in Fig. \ref{fig:architecture}, the design of TSE-Net follows the self-training pipeline, with two innovations specific to the monocular height estimation task. First, an additional network, termed the exam network, accumulates the weights of the student network via exponential moving average (EMA) updates to maintain temporally stabilized performance. Second, the teacher network is designed as a multi-task architecture with an additional classification branch, which mitigates the long-tailed distribution problem and provides probabilities that serve as proxies for pseudo-label filtering.
\subsection{Teacher-Student-Exam Self-training}
Following the self-training paradigm, the pipeline includes two primary networks: the teacher and student. Both networks are trained simultaneously on the labeled dataset, while during training they are also exposed to unlabeled data in two different views, strong augmentations for the student network, and weak ones for the teacher. The teacher network predicts pseudo-labels for the unlabeled samples to supervise the student network, thus allowing the student to learn from data with greater diversity.

Conventionally, the teacher network maintains a temporal ensemble of the student network via EMA updates. However, our experiments indicate that when the teacher network differs slightly in architecture and objectives from the student, it may not represent the optimal state of the student. Therefore, we introduce an exam network that maintains a temporal ensemble of the student while sharing the same architecture and objective. The exam network is updated only as the EMA of the student's weights and is not updated through gradient backpropagation.

To generate two different data views for training, weak and strong augmentations are applied to unlabeled data samples. The weak augmentations involve subtle geometric transformations that do not alter scale or spectral patterns, including random flipping and 90-degree rotations. The strong augmentations include geometric and spectral transformations, such as random rotations with arbitrary angles, random cropping to half size, random gamma, brightness, and contrast adjustments, and random Gaussian blurring.
\subsection{Multi-task Network as Teacher}
The teacher network is designed as a multi-task model comprising a classification and a regression branch. The regression branch predicts pixel-wise height values, while the classification branch predicts pixel-wise height classes defined by the hierarchical bi-cut. This multi-task design serves two purposes: first, the classification branch provides per-pixel class probabilities that are used for pseudo-label filtering. Second, discretizing the continuous output space in a balanced way mitigates biases caused by the long-tailed data distribution.
\subsubsection{Architecture}
The multi-task network adopts an encoder-decoder structure. The encoder processes the input image to extract condensed feature representations, which are then upsampled by the decoder to match the input image. The resulting feature tensor is fed into a regression output layer to predict height values and transformed by a linear layer into classification features, which are used by the classification output layer to predict height classes for each pixel.
\subsubsection{Hierarchical Bi-Cut for Classification Branch}
\begin{figure}[!htbp]
    \centering
    \includegraphics[width=\linewidth]{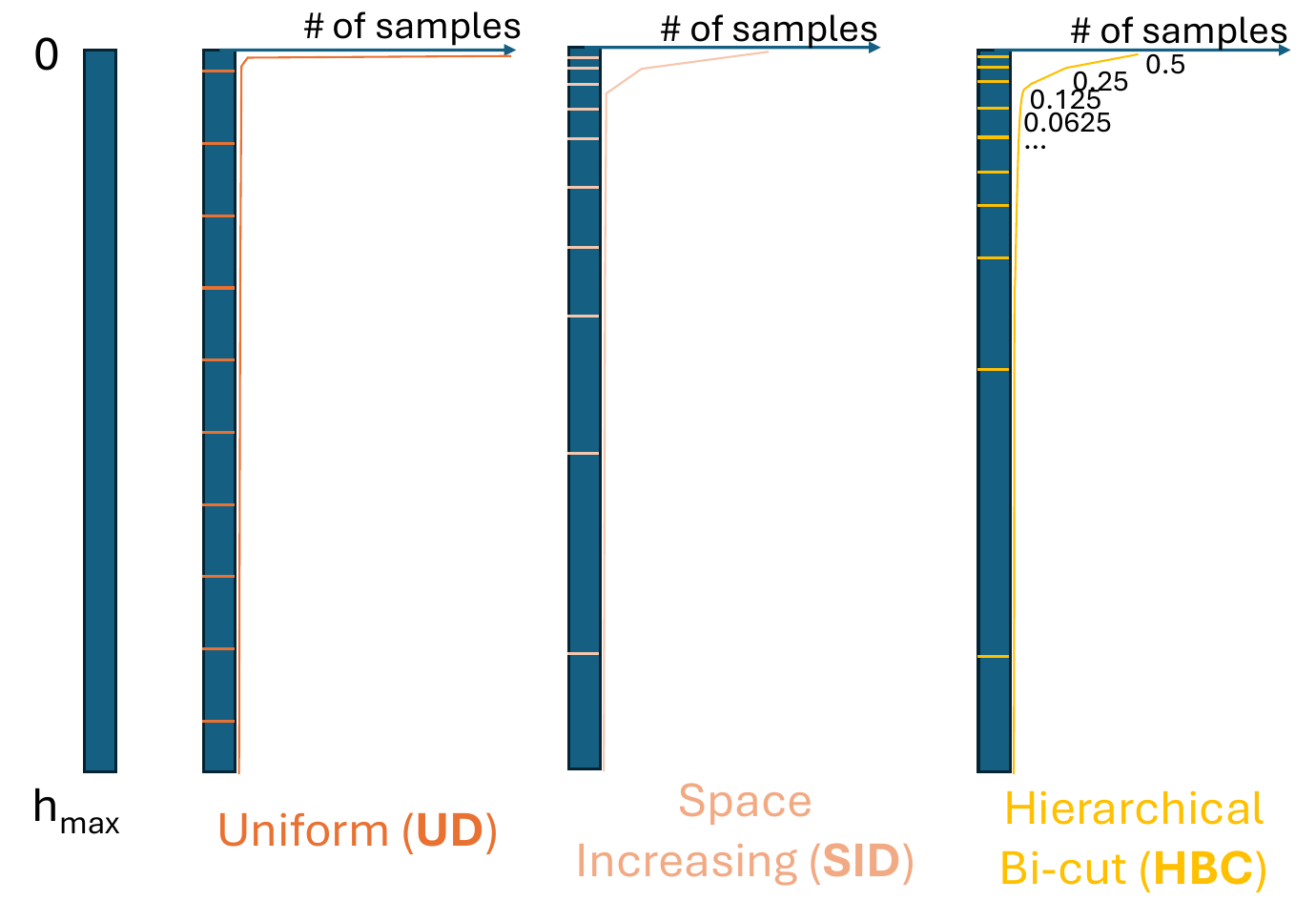}
    \caption{Different discretization strategies for converting regression to classification. $h_\text{max}$: maximal height value in the dataset. Uniform discretization (UD): divides the height range uniformly. Space-increasing discretization (SID): divides the height range logarithmically. Hierarchical Bi-cut (HBC): divides the height range hierarchically so that each cut splits the samples in half.}
    \label{fig:bins}
\end{figure}
To transform the regression problem into a classification one, the continuous output space $H=(h_\text{min}, h_\text{max})$ is discretized into $N$ bins. Previous works on monocular depth and height estimation employed three main strategies (see Fig. \ref{fig:bins}):
\begin{itemize}
    \item \textbf{Uniform Discretization (UD)}: Divides the height range evenly: The lower bin edge for class $i$ is 
    \begin{equation}
        e_i=h_\text{min}+\displaystyle\frac{h_\text{max}-h_\text{min}}{N}i.
    \end{equation}
    \item \textbf{Space-Increasing Discretization (SID)} \cite{fu2018a}, also referred to as \textbf{Log-Uniform Discretization} in \cite{bhat2022}: Applies uniform division in logarithm space to have increased intervals of large height/depth values: The lower bin edge for class $i$ is written as
    \begin{equation}
        e_i=\displaystyle\exp(\displaystyle\log h_\text{min} + \displaystyle\frac{\log h_\text{max} -\log h_\text{min}}{N}i).
    \end{equation}
    \item \textbf{Adaptive Bins (AdaBins)} \cite{bhat2022}: Learns bin edges adaptively for each image according to its height value distribution.
\end{itemize}

\begin{figure}[!htbp]
    \centering
    \includegraphics[width=\linewidth]{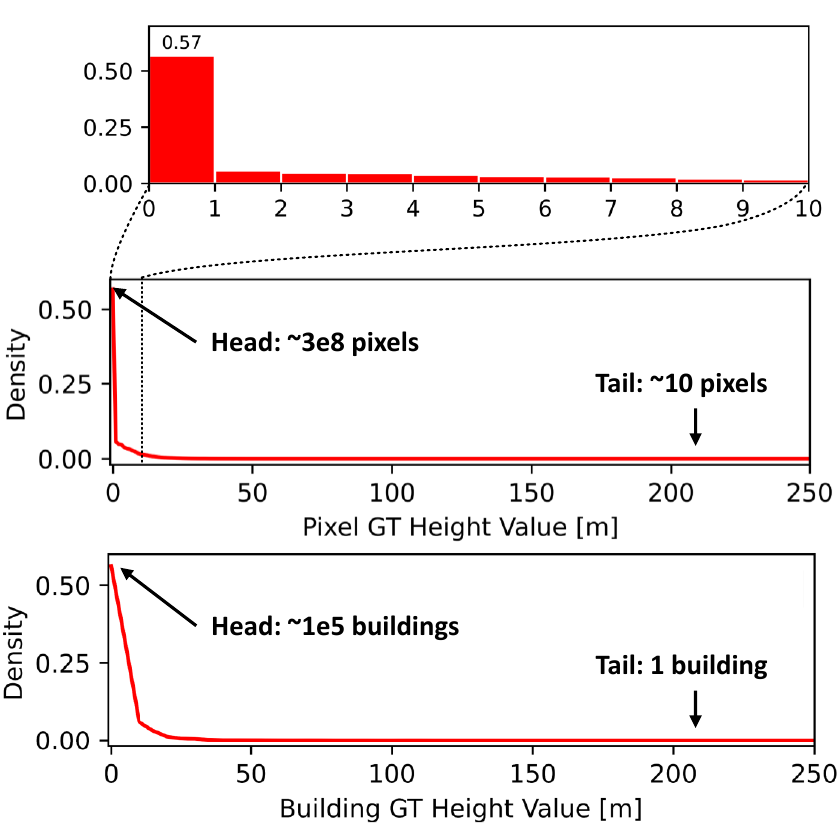}
    \caption{Height value Distribution of the GBH training and validation Sets. The distribution exhibits a pronounced long-tailed pattern: approximately 3e8 pixels (57 \% of the total) fall within background regions below 1 m, while the frequency of pixels at larger heights decreases sharply to about 10 pixels per 1 m bin. A similar long-tailed distribution is observed for the building ground-truth height values.}
    \label{fig:distribution}
\end{figure}

Unlike depth values in monocular depth estimation, height values in monocular height estimation are extremely long-tailed \cite{chen2023g} (see Fig. \ref{fig:distribution}). For example, in the GBH dataset, pixels lower than 1 m account for over half of all pixels, whereas high-value pixels are sparse. Improper class definitions can thus lead to strong model bias and degrade classification accuracy. From this perspective, UD fails to consider the data distribution. More specifically, when the number of classes is relatively small, the first class will encompass almost all the pixels, which makes the classification meaningless. SID partly addresses the issue, but lacks adaptability to the dataset. AdaBins adapts to each image, but is computationally more complex, and the classes across images are not consistent, which poses challenges when filtering the pseudo-labels based on the class probabilities. Furthermore, under semi-supervised learning settings where labeled samples are limited, some classes defined by UD or SID may never appear during training.

To overcome these limitations, we proposed Hierarchical Bi-Cut (HBC), which recursively divides the height range into halves. The class edges, termed HBC-points, are defined iteratively. The first HBC-point divides the full range at the median height value, separating height values for class $0$ and the remaining classes. Subsequent cuts divide the remaining upper ranges by their medians until all classes from $1$ to $N-1$ are defined. Formally, the $i$-th HBC-point $Q_i$, separating class $i$ from higher classes, is defined as
\begin{equation}
    Q_i: P(h<Q_i)=1-(\frac{1}{2})^{i+1},
\end{equation}
where h is the pixel height value. In practice, the HBC-points are defined for a dataset once by a binary search. This hierarchical approach balances each binary split between the lower class and the remaining higher classes, addressing the imbalance problem without adding computational overhead.

Following HBC, the imbalanced classification problem is decomposed into $N-1$ binary classification tasks, naturally aligning with ordinal regression \cite{herbrich1999,harrell2015a}. For each binary classification problem, one has to decide whether a pixel belongs to this class or the following classes with higher values. Intuitively, a pixel belonging to classes with higher values than class $c+1$ should bear a height value also higher than class $c$.

To implement the intuition, ordinal regression aims to learn the objective of multi-hot labels. Different from one-hot labels, where only the ground truth class itself is expected to be positive, the multi-hot labels set all the classes equal to or lower than the ground truth class as positive. That is,
\begin{equation}
    y = \displaystyle[\underbrace{1, 1, \cdots, 1}_{c}, \underbrace{0, \cdots, 0, 0}_{N-c}],
\end{equation}
where $y$ is the label for one pixel belonging to class $c$. The ordinal regression loss is computed as
\begin{equation}
    \label{eqn:cls_loss}
    L_{cls}=\frac{1}{N-1}\sum_{i=0}^{N-1}(-y_i\log p_i-(1-y_i)\log(1-p_i)),
\end{equation}
where $y_i$ is the $i$-th element of the multi-hot label $y$, and $p_i$ is the $i$-th element of the predicted probabilities $p$.

For pseudo-label filtering, the class probability of class $i$ is computed as
\begin{equation}
    \label{eqn:cls_prob_comp}
    q_i =
    \begin{cases}
    1 - p_0, & i = 0, \\[6pt]
    (1 - p_i)\displaystyle\prod_{c=0}^{i-1} p_c, & 1 \le i \le N - 2, \\[6pt]
    \displaystyle\prod_{c=0}^{N-1} p_c, & i = N - 1.
    \end{cases}
\end{equation}
\subsubsection{Regression Branch}
The regression branch is supervised using the L1 loss. For one pixel, that is
\begin{equation}
    \label{eqn:reg_loss}
    \mathcal{L}_{reg}=|\hat{h}-h|.
\end{equation}
where $\hat{h}$ is the predicted height value, and $h$ is the ground truth height value.
\subsubsection{Alignment of Classification Probabilities via Plackett-Luce Model}
Since class probabilities are used for pseudo-label filtering, they should ideally correlate with pseudo-label accuracy: larger regression errors should correspond to lower class probabilities. To this end, we employ a Plackett-Luce (PL) model \cite{plackett1975,luce2005} to enforce ranking consistency between classification probabilities and pseudo-label errors.

The PL model is a model for ranking data working with listwise preferences. Given a list of $M$ pixels from the pseudo-labels predicted by the teacher network, their class probabilities are denoted as 
\begin{equation}
    \mathbf{q}=\{q_0,q_1,\cdots,q_{M-1}\},
\end{equation}
and their pseudo-label absolute errors are denoted as
\begin{equation}
    \mathbf{\epsilon}=\{\epsilon_0,\epsilon_1,\cdots,\epsilon_{M-1}\},
\end{equation}
where 
\begin{equation}
    \epsilon_j=|h_j-\hat{h}_j|
\end{equation}
is the absolute error of pixel $j$, that is, the absolute difference between the ground truth and predicted height values of pixel $j$, $h_j$ and $\hat{h}_j$, respectively. The PL model is designed to align the ranking of $\mathbf{q}$ with that of $\mathbf{\epsilon}$.

The process is achieved by a loss function representing the negative log-likelihood of a random permutation of $\mathbf{q}$. The probability of a permutation $\pi$ of $\arg \mathbf{q}$ is given by
\begin{equation}
    P(\pi \mid \boldsymbol{q}) =
    \prod_{j=0}^{M-2}
    \frac{q_{\pi_j}}{\sum_{k=j}^{M-1} q_{\pi_k}}.
\end{equation}
Given the ground truth permutation $\Pi$ which reflects the ranking of $\mathbf{\epsilon}$, the probability is written as
\begin{equation}
    P(\Pi \mid \boldsymbol{q}) =
    \prod_{j=0}^{M-2}
    \frac{q_{\Pi_j}}{\sum_{k=j}^{M-1} q_{\Pi_k}}.
\end{equation}
The PL model is then optimized with the loss function written as the negative log-likelihood of the ground truth permutation, that is
\begin{equation}
    \label{eqn:pl_loss}
    \mathcal{L}_{\mathrm{PL}} = -\log P(\Pi \mid \boldsymbol{q}).
\end{equation}
\subsubsection{Pseudo-Label Filtering from Ranking}
Using the class probabilities, regression pseudo-labels are filtered according to their ranking. For a batch of $B$ predicted height maps and class probability maps $\hat{\mathcal{Q}}$, the pixels are sorted by their class probabilities, yielding a normalized ranking map
\begin{equation}
    \mathcal{R}=\frac{\text{argsort} \hat{\mathcal{Q}}}{|\hat{\mathcal{Q}}|},
\end{equation}
where $|\hat{\mathcal{Q}}|$ denotes the total number of pixels in $\hat{\mathcal{Q}}$. Pseudo-labels are retained for supervising the student network if their normalized ranking exceeds a ranking threshold $r$. The pseudo-label filtering mask $\mathcal{M}$ is then defined as
\begin{equation}
    \mathcal{M}_{j}= 1 ~\text{if}~ \mathcal{R}_j>r ~\text{else}~ 0,
\end{equation}
where $j$ is the pixel index.

The ranking threshold decays from its initial value 1 according to a factor $\lambda$ during training, that is
\begin{equation}
    r_{t+1}=\max(r_t \lambda, 0.5).
\end{equation}
This ensures that unlabeled samples are initially excluded when the network is not well-trained and predictions are uncertain. As training progresses, the threshold decreases, gradually incorporating more unlabeled samples into the training pool. This progressive inclusion prevents network collapse caused by unreliable pseudo-labels.
\subsubsection{Loss Functions}
The teacher network is trained with a combined loss, consisting of the classification loss (see Eqn. \ref{eqn:cls_loss}), regression loss (see Eqn. \ref{eqn:reg_loss}), and PL model loss (see Eqn. \ref{eqn:pl_loss}), written as
\begin{equation}
\mathcal{L}_\text{sup}^{T}=\mathcal{L}_{cls}+\mathcal{L}_{reg}^{T}+\mathcal{L}_{PL}
\end{equation}
\subsection{Regression Networks as Student and Exam}
The student network expands the model's capacity to handle data with greater variability and is supervised by the filtered pseudo-labels produced by the teacher network. Given pseudo-labels $\tilde{h}^T$ from the teacher and the filtering mask $\mathcal{M}$ derived from ranking thresholding, the student network is trained via 
\begin{equation}
    \mathcal{L}_\text{unlabeled}=\mathcal{L}_{reg}^{S}\odot \mathcal{M}=|\hat{h}^{S}-\tilde{h}^T|\odot \mathcal{M},
\end{equation}
where $\hat{h}^{S}$ represents the predictions of the student network on unlabeled data samples.

The exam network does not participate in loss computation or gradient backpropagation. Its weights are updated exclusively as the EMA of the student network. During inference and testing, the exam network is used to generate the final outputs.
\vspace{1em}

In summary, the training procedure of TSE-Net is outlined in Algorithm \ref{alg:training}.

\begin{algorithm*}[!htbp]
\caption{Training Step of TSE-Net}
\label{alg:training}
\begin{algorithmic}[1]
\Require Image $x$, ground-truth nDSM $y$, unlabeled image $u$

\State $(\hat{h}^{T}, p^{T}) \gets \text{Teacher}(x)$ 
\Comment{Teacher predictions (height and class)}

\State $\hat{h}^{S} \gets \text{Student}(x)$ 
\Comment{Student prediction (height)}

\State $q^{T} \gets \text{Binary2Class}(p^{T})$

\Comment{Convert binary probabilities into class probabilities}

\State $\mathcal{L}_{\text{sup}}^{T} \gets \mathcal{L}{\text{cls}}(p^{T}, \text{ndsm2cls}(h)) + \mathcal{L}_{\text{reg}}(\hat{h}^{T}, h) + \mathcal{L}_{\text{PL}}(q^{T}, \text{L1}(\hat{h}^T, h))$

\State $\mathcal{L}_{\text{sup}}^{S} \gets \mathcal{L}_{\text{reg}}(\hat{h}^{S}, h)$

\Statex   

\State $v_{1} \gets \text{AugmentWeak}(u)$

\State $(\tilde{h}_{1}^T, \tilde{p}_{1}^T) \gets \text{Teacher}(v_{1})$ 
\Comment{Pseudo-labels}

\State $(v_{2}, \tilde{h}_{2}^T, \tilde{p}_{2}^T) \gets \text{AugmentStrong}(v_{1}, \tilde{h}_{1}^T, \tilde{p}_{1}^T)$

\State $\tilde{q}_2^{T} \gets \text{Binary2Class}(\tilde{p}_2^{T})$

\State $\mathcal{M} \gets \text{rank}(\tilde{q}_{2}^T) > \tau_{\text{rank}}$ 
\Comment{Pseudo-label mask}

\State $\hat{h}_{2}^{S} \gets \text{Student}(v_{2})$

\State $\mathcal{L}_{\text{unlabeled}} \gets \mathcal{L}_{\text{reg}}^{S}(\hat{h}_{2}^{S}, \tilde{h}_{2}^T) \odot \mathcal{M}$

\Statex

\State $\mathcal{L} \gets \mathcal{L}_{\text{sup}}^{T} + \mathcal{L}_{\text{sup}}^{S} + \mathcal{L}_{\text{unlabeled}}$

\State Backpropagate $\mathcal{L}$

\State Update(Student) \Comment{Adam optimizer}

\State EMA\_Update(Exam, Student) 
\Comment{Exponential moving average}
\end{algorithmic}
\end{algorithm*}

\section{Experiments}\label{chap:experiments}
\subsection{Datasets}
\begin{table*}[!htbp]
    \centering
    \footnotesize
    \caption{Overview of the Datasets Used for the Experiments.}
    \resizebox{\textwidth}{!}{%
    \begin{tabular}{cccc}
    \toprule
    & ISPRS Vaihingen \cite{isprsv,lesaux2019} & SynRS3D \cite{song2024} & GBH \\ \midrule
    Sensor & Airborne & Synthetic & Planet \\ \midrule
    Involved Areas & Vaihingen an der Enz, Germany & n.a. & 18 cities around the globe \\ \midrule
    Resolution (m/pixel) & 0.09 & 0.6--1 & 5 (upsampled to 3) \\ \midrule
    \multirow{3}{*}{Semantic Classes} & Impervious, Building, & Bareland, Rangeland, & Building, \\
    & Low Vegetation, & Developed Space, Road, Tree, & Non-building\\
    & Tree, Car & Water, Agriculture Land, Building \\ \midrule
    Original Image Size & $\sim$ 2000 $\times$ 2000 & 512 $\times$ 512 & 256 $\times$ 256 \\ \midrule
    Cropped Image Size & 256 $\times$ 256 & 512 $\times$ 512 & 256 $\times$ 256 \\ \midrule
    Number of Cropped Images & 2,838 & 5516 & 8703 \\ \midrule
    Training: Validation: Test & 1060:264:1514 (4:1:5) & 3862:551:1103 (7:1:2) & 5442:1477:1784 (6:2:2)\\ \midrule
    \multirow{5}{*}{Labeled: Unlabeled in Training} & 1:1059 (0.1\%) & 3:3859 (0.1\%) & 5:5437 (0.1\%) \\
    & 5:1055 (0.5\%) & 19:3843 (0.5\%) & 27:5415 (0.5\%) \\
    & 10:1050 (1\%) & 38:3824 (1\%) & 54:5388 (1\%) \\
    & 53:1007 (5\%) & 190:3672 (5\%) & 272:5170 (5\%) \\
    & 106:954 (10\%) & 380:3482 (10\%) & 544:4898 (10\%) \\ \bottomrule
    \end{tabular}}
    \label{tab:data}
\end{table*}
To validate the proposed method, three datasets are used in experiments, namely ISPRS Vaihingen, SynRS3D and GBH. These datasets cover different domains (synthetic and real), sensors (infrared and RGB), platforms (aerial and space), and spatial resolutions (from 0.09 m to 3 m). A detailed description of the datasets is provided in Table \ref{tab:data}. Note that the ISPRS Vaihingen dataset is split so that the test set is consistent with the setting of the Data Fusion Contest. Additionally, it simulates extreme scenarios in which labeled training data are limited.
\subsection{Evaluation Metrics}
The methods are evaluated in terms of root mean square errors (RMSEs), computed over all pixels and across different land cover classes. To assess the models' ability to mitigate long-tailed biases, an averaged RMSE of building heights across the full height range is also calculated. For this purpose, buildings are assumed to have Level of Detail 1 (LoD1), where a single height value is assigned to each building instance. The predicted and ground truth building heights are then defined as the median height value within each building footprint coverage. Additionally, the mean relative error is computed for each building instance to more accurately reflect the relative scale of prediction errors.

\subsection{Implementation Details}
U-Net \cite{ronneberger2015} is set as the backbone network. The experiments are designed as follows:
\begin{itemize}
    \item The backbone network is trained with all available data (100\%), serving as an upper-bound reference for performance.
    \item The backbones are trained in a fully supervised manner with 0.1\%, 0.5\%, 1\%, 5\%, and 10\% of the labeled data, providing baselines for varying levels of label scarcity.
    \item The proposed networks are trained in a semi-supervised manner using the same labeled subsets (0.1\%--10\%) together with the remaining unlabeled images, and the results are compared against the baselines.
\end{itemize}

All networks are trained using the Adam optimizer with a learning rate of 1e-4 for 200 epochs. The epoch yielding the best performance on the validation set is selected for testing. Each experiment is repeated three times, and the reported metrics correspond to the average across these runs.
\section{Results}\label{chap:results}
\begin{figure*}[!htbp]
    \centering
    \includegraphics[width=\linewidth]{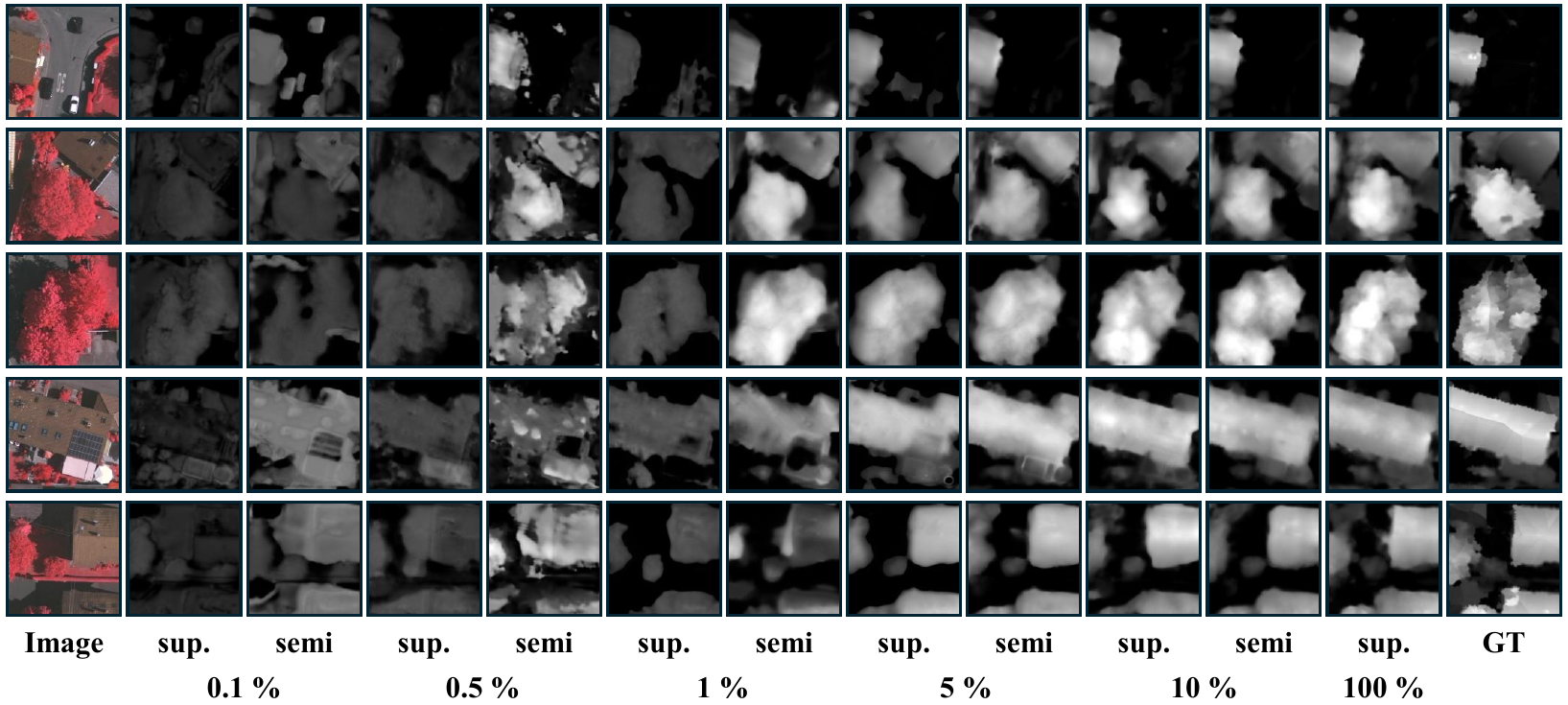}
    \caption{Qualitative results on ISPRS Vaihingen. sup.: supervised learning; semi: semi-supervised learning; GT: ground truth; percentage on the bottom denotes the labeled ratio.}
    \label{fig:vai_res}
\end{figure*}

\begin{figure*}[!htbp]
    \centering
    \includegraphics[width=\linewidth]{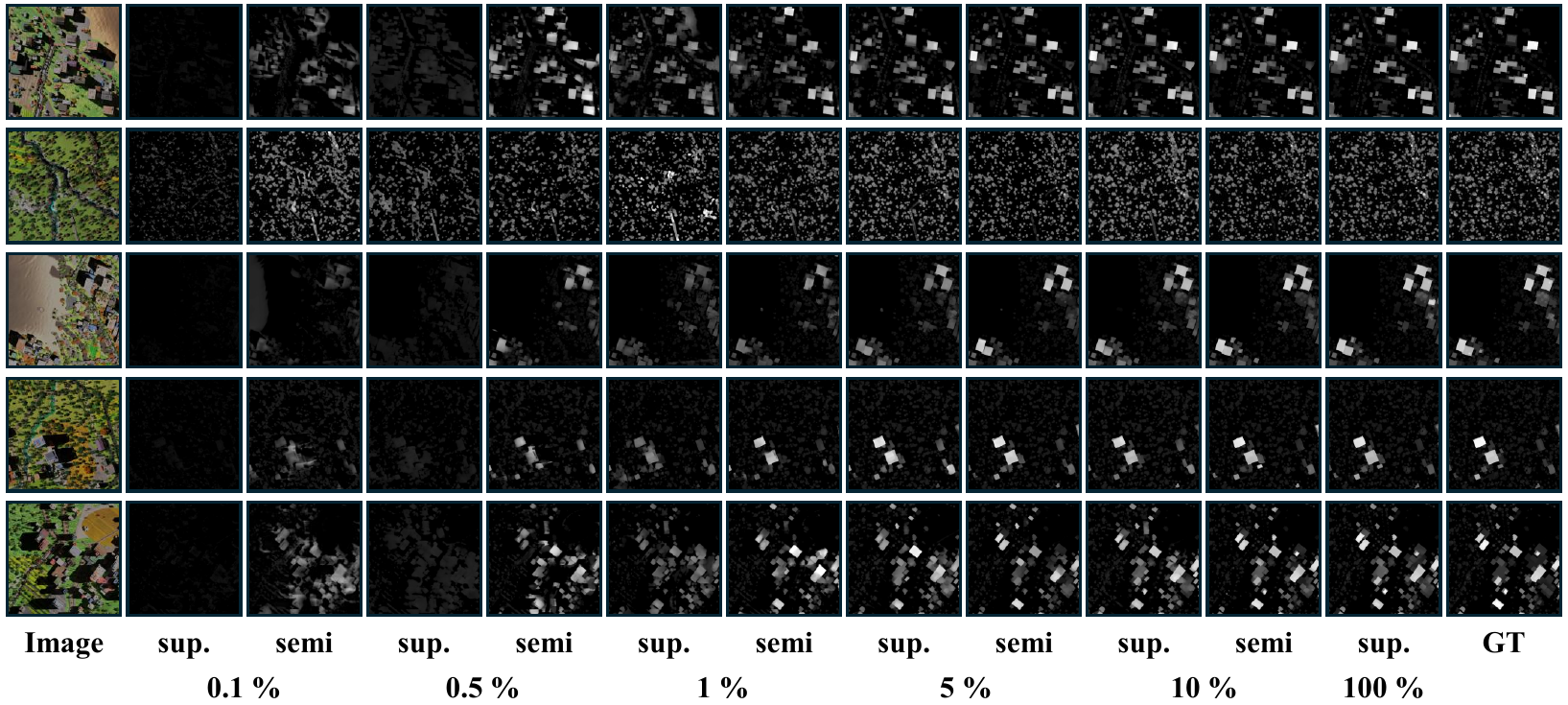}
    \caption{Qualitative results on SynRS3D. sup.: supervised learning; semi: semi-supervised learning; GT: ground truth; percentage on the bottom denotes the labeled ratio.}
    \label{fig:syn_res}
\end{figure*}

\begin{figure*}[!htbp]
    \centering
    \includegraphics[width=\linewidth]{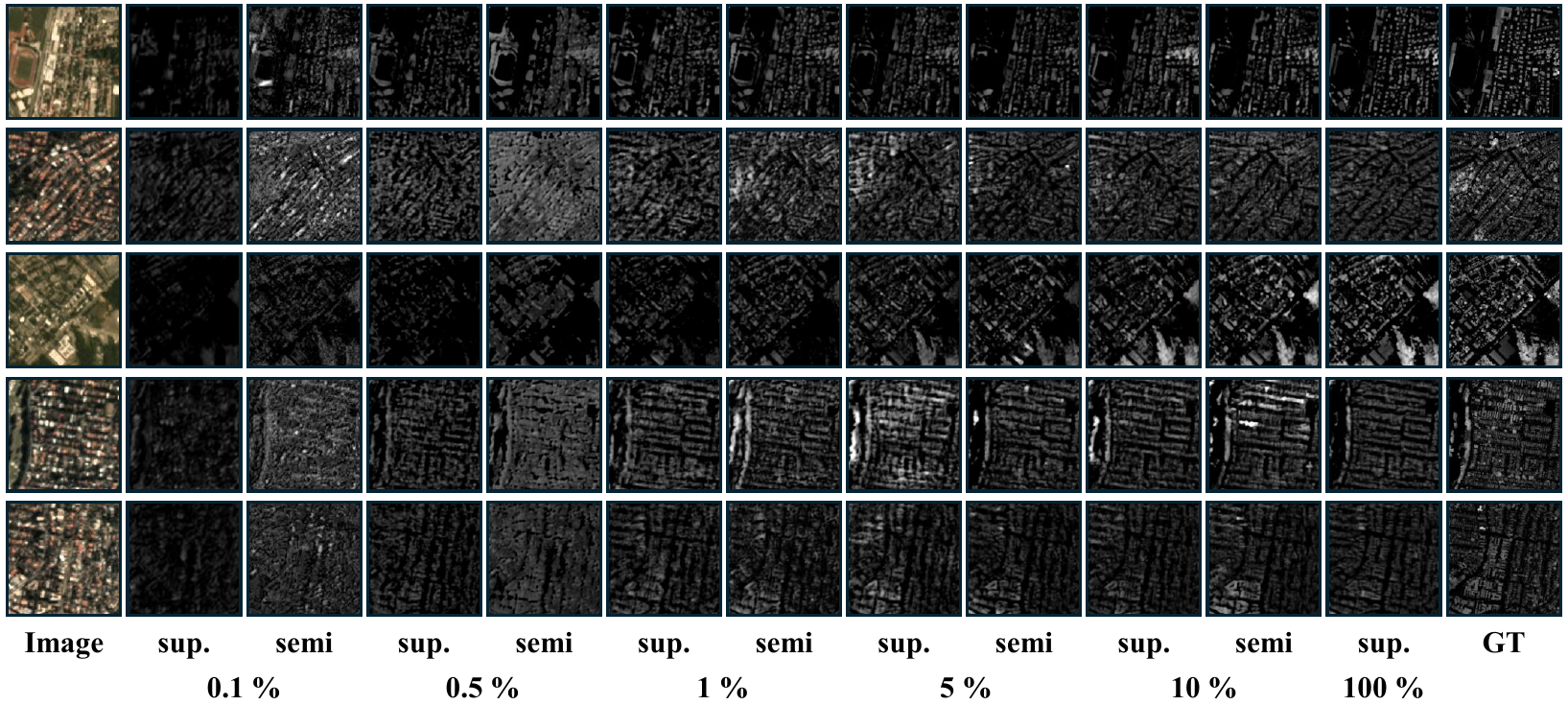}
    \caption{Qualitative results on GBH. sup.: supervised learning; semi: semi-supervised learning; GT: ground truth; percentage on the bottom denotes the labeled ratio.}
    \label{fig:gbh_res}
\end{figure*}

The quantitative results are reported in Table \ref{tab:vai_res}, Table \ref{tab:syn_res}, and Table \ref{tab:gbh_res}, with corresponding qualitative examples shown in Fig. \ref{fig:vai_res}, Fig. \ref{fig:syn_res}, and Fig. \ref{fig:gbh_res}.

Overall, the results confirm that performance improves as the proportion of labeled data increases. First, with more labels, the total RMSEs and relative building errors decrease across all three datasets. Second, across land cover types, those characterized by lower heights, such as impervious surfaces, low vegetation, and cars in ISPRS Vaihingen (Table \ref{tab:vai_res}), or bareland, rangeland, developed space, roads, water, and agricultural land in SynRS3D (Table \ref{tab:syn_res}), often show limited or even degraded improvement as the label proportion increases. In contrast, tall classes such as buildings and trees benefit more consistently. Third, given the long-tailed distribution of building heights, adding more labels also exposes the network to extreme values, which improves the balanced building RMSEs across all three datasets.

\begin{table*}[!htbp]
    \centering
    \footnotesize
    \caption{Experiment Results on ISPRS Vaihingen. The RMSEs are reported in meters for all pixels (Total), for each land cover type, and for buildings averaged by different heights (Building Balanced). Building Relative denotes the mean relative error of building instances.}
    \resizebox{\textwidth}{!}{%
    \begin{tabular}{c|c|c|ccccc|cc}
    \toprule
         Labeled & Mode & Total & Impervious & Building & Low Vegetation & Tree & Car & \multicolumn{2}{c}{Building} \\
         Percentage & & & & & & & & Balanced & Relative \\ \midrule
         \multirow{2}{*}{0.1} & sup. & 4.0747 & 1.4568 & 6.4593 & 1.9049 & 4.4689 & 1.5476 & 8.5296 & 0.8161 \\
         & semi & 3.7562 & 2.2343 & 4.9890 & 2.1756 & 4.7172 & 3.6533 & 6.9682 & 0.4565 \\ \midrule \midrule
         \multirow{2}{*}{0.5} & sup. & 3.9268 & 1.5017 & 6.1442 & 1.8632 & 4.3729 & 1.8814 & 8.3209 & 0.7389  \\
         & semi & 3.6847 & 2.0470 & 5.0112 & 3.1277 & 3.9757 & 2.4542 & 7.0436 & 0.5418 \\ \midrule \midrule
         \multirow{2}{*}{1} & sup. & 3.7425 & 1.4289 & 5.8168 & 1.8083 & 4.2111 & 1.6183 & 8.0985 & 0.6915 \\
         & semi & 3.4485 & 1.5363 & 5.0094 & 2.8263 & 3.6228 & 1.9090 & 7.1249 & 0.5495 \\ \midrule \midrule
         \multirow{2}{*}{5} & sup. & 2.8503 & 1.3454 & 4.0585 & 2.1265 & 3.1928 & 1.3118 & 6.0352 & 0.3925 \\
         & semi & 2.7620 & 1.3847 & 3.9119 & 2.3412 & 2.8683 & 1.4005 & 5.3548 & 0.3343 \\ \midrule \midrule
         \multirow{2}{*}{10} & sup. & 2.4505 & 1.4451 & 3.3051 & 1.8496 & 2.7554 & 1.6840 & 4.7989 & 0.3038 \\
         & semi & 2.3235 & 1.4149 & 3.2383 & 1.5097 & 2.6247 & 1.3235 & 4.7284 & 0.2991 \\ \midrule \midrule
         100 & sup. & 1.9210 & 1.0131 & 2.7796 & 1.2026 & 2.1381 & 0.9409 & 3.9035 & 0.2432 \\ \bottomrule
    \end{tabular}}
    \label{tab:vai_res}
\end{table*}
\begin{table*}[!htbp]
    \centering
    \footnotesize
    \caption{Experiment Results on SynRS3D. The RMSEs are reported in meters for all pixels (Total), for each land cover type, and for buildings averaged by different heights (Building Balanced). Building Relative denotes the mean relative error of building instances.}
    \resizebox{\textwidth}{!}{%
    \begin{tabular}{c|c|c|cccccccc|cc}
    \toprule
         Labeled & Mode & Total & Bareland & Rangeland & Developed & Road & Tree & Water & Agriculture & Building & \multicolumn{2}{c}{Building} \\
         Percentage & & & & & Space & & & & Land & & Balanced & Relative \\ \midrule
         \multirow{2}{*}{0.1} & sup. & 22.0667 & 0.9753 & 0.6789 & 1.8512 & 2.3424 & 8.7109 & 1.4581 & 0.8976 & 52.3291 & 49.1640 & 0.8973 \\
         & semi & 17.8707 & 5.7502 & 3.8579 & 13.9634 & 8.7156 & 7.6223 & 5.9271 & 4.1192 & 39.8991 & 39.7101 & 0.8879 \\ \midrule \midrule
         \multirow{2}{*}{0.5} & sup. & 19.8165 & 4.1891 & 2.6496 & 6.2870 & 4.9685 & 6.4080 & 3.6529 & 2.8220 & 46.4933 & 43.7314 & 0.7198 \\
         & semi & 15.4076 & 5.4603 & 3.1961 & 14.8628 & 7.2410 & 6.4711 & 5.3315 & 5.0473 & 33.5954 & 36.3823 & 0.8140 \\ \midrule \midrule
         \multirow{2}{*}{1} & sup. & 15.6627 & 5.9417 & 2.9399 & 11.8148 & 6.9981 & 6.3581 & 5.5319 & 3.8327 & 35.0818 & 36.2907 & 0.8467 \\
         & semi & 14.1046 & 4.6899 & 2.9150 & 10.7695 & 4.5643 & 5.6551 & 3.7841 & 3.4999 & 31.7323 & 35.0314 & 0.7961 \\ \midrule \midrule
         \multirow{2}{*}{5} & sup. & 12.0842 & 2.9985 & 2.6244 & 8.8715 & 3.8812 & 5.0232 & 2.5368 & 2.3090 & 27.3505 & 32.5352 & 0.7804 \\
         & semi & 10.8257 & 2.0875 & 2.0208 & 5.9653 & 2.7044 & 4.6858 & 1.5741 & 1.7733 & 24.9419 & 32.2449 & 0.6752 \\ \midrule \midrule
         \multirow{2}{*}{10} & sup. & 11.0390 & 2.5044 & 2.2594 & 6.9351 & 3.3501 & 4.5911 & 2.1931 & 2.1058 & 25.2400 & 32.0167 & 0.7106 \\
         & semi & 9.1434 & 1.8249 & 1.9821 & 4.7824 & 2.5746 & 4.2950 & 1.4224 & 1.4336 & 20.9870 & 30.6148 & 0.6173 \\ \midrule \midrule
         100 & sup. & 7.6350 & 1.6509 & 1.8807 & 4.2640 & 2.2845 & 3.7911 & 1.2005 & 1.2886 & 17.4010 & 30.1244 & 0.5691 \\ \bottomrule
    \end{tabular}}
    \label{tab:syn_res}
\end{table*}
\begin{table*}[!htbp]
    \centering
    \footnotesize
    \caption{Experiment Results on GBH. The RMSEs are reported in meters for all pixels (Total), for each land cover type, and for buildings averaged by different heights (Building Balanced). Building Relative denotes the mean relative error of building instances.}
    \begin{tabular}{c|c|c|cc|cc}
    \toprule
         Labeled & Mode & Total & Non-building & Building & \multicolumn{2}{c}{Building} \\
         Percentage & & & & & Balanced & Relative \\ \midrule
         \multirow{2}{*}{0.1} & sup. & 7.2996 & 4.8074 & 13.2922 & 47.0171 & 0.7981 \\
         & semi & 7.1086 & 5.0194 & 12.4127 & 46.2821 & 0.5932 \\ \midrule \midrule
         \multirow{2}{*}{0.5} & sup. & 7.0267 & 4.6523 & 12.7587 & 47.7627 & 0.7350 \\
         & semi & 6.7179 & 4.8258 & 11.5936 & 45.6716 & 0.5258 \\ \midrule \midrule
         \multirow{2}{*}{1} & sup. & 6.5333 & 4.6039 & 11.4264 & 45.6464 & 0.5984 \\
         & semi & 6.2725 & 4.8843 & 10.1108 & 42.4920 & 0.5078 \\ \midrule \midrule
         \multirow{2}{*}{5} & sup. & 5.8077 & 4.4457 & 9.5237 & 40.2960 & 0.4878 \\
         & semi & 5.7340 & 4.1448 & 9.8532 & 40.8399 & 0.5022 \\ \midrule \midrule
         \multirow{2}{*}{10} & sup. & 5.5087 & 4.1569 & 9.1510 & 40.1568 & 0.4703 \\
         & semi & 5.3971 & 4.0124 & 9.0781 & 40.4248 & 0.4704 \\ \midrule \midrule
         100 & sup. & 4.8920 & 3.8018 & 7.9179 & 37.4738 & 0.3845 \\ \bottomrule
    \end{tabular}
    \label{tab:gbh_res}
\end{table*}

When unlabeled data are incorporated through the self-training pipeline, three main effects become evident, and all three suggest that unlabeled data are effectively utilized in a similar manner to labeled data. This is because the same behaviors can also be observed when the proportion of labeled data is increased, as discussed above.

First, semi-supervised learning yields substantial overall improvements, particularly at very low label proportions (0.1\%--1\%). For example, on the ISPRS Vaihingen dataset, the total RMSE drops from 4.0747 (supervised, 0.1\%) to 3.7562 (semi-supervised, 0.1\%), which corresponds to filling 14.79\% of the gap towards the 100\% fully supervised model. Similarly, the gaps closed on SynRS3D and GBH with only 0.1\% labeled data are 29.07\% and 7.93\%, respectively. Furthermore, on SynRS3D (Table \ref{tab:syn_res}), the semi-supervised model trained with 0.1\% labeled data outperforms the supervised model trained with 0.5\%, and the semi-supervised model trained with 0.5\% labeled data outperforms the supervised model trained with 1\%. The improvements are also demonstrated by the relative building errors. In addition, these gains are reflected qualitatively: with only 0.1\% labeled data, supervised predictions show no discernible structure, whereas semi-supervised predictions capture spatial context and object layout. At 0.5\%--1\%, supervised models start producing meaningful patterns, but semi-supervised learning further mitigates underestimation and pushes predictions closer to the ground truth.

Second, the use of unlabeled data amplifies errors for low-height classes while improving predictions for tall classes, mirroring the outcomes observed with full supervision on larger proportions of labeled data. In ISPRS Vaihingen (Table \ref{tab:vai_res}), RMSEs for impervious surfaces and cars increase under semi-supervised learning, while buildings and trees improve significantly. The same pattern holds in SynRS3D (Table \ref{tab:syn_res}), where semi-supervised models yield higher errors for rangeland and agriculture but substantially reduce errors for roads and buildings. GBH (Table \ref{tab:gbh_res}) shows a similar contrast: non-building pixels remain stable or slightly worse, while building RMSEs improve notably, especially at 0.5 \% and 1 \%. The consistency of this class-dependent trade-off across all datasets underscores that unlabeled data reinforce similar learning behaviors to those driven by labeled data.

Third, unlabeled data mitigate the bias caused by the long-tailed distribution of building heights, improving balanced building RMSEs. On ISPRS Vaihingen, for instance, the balanced building RMSE decreases from 8.5296 (supervised, 0.1\%) to 6.9682 (semi-supervised, 0.1\%), an 18.31\% improvement (Table \ref{tab:vai_res}). On SynRS3D, the improvement is also clear, with the balanced building RMSE dropping from 49.1640 (supervised, 0.1\%) to 39.7101 (semi-supervised, 0.1\%), a 19.23\% improvement (Table \ref{tab:syn_res}). Even on GBH, where the total RMSE gain is less pronounced, the balanced building RMSE improves consistently at low label proportions (Table \ref{tab:gbh_res}). Since similar de-biasing effects are observed when more labeled data are added, these results again confirm that unlabeled data extend the training signal in a manner consistent with labeled data.
\section{Ablation Studies}\label{chap:discussions}
\begin{table*}[!htbp]
    \centering
    \footnotesize
    \caption{Ablation Studies on ISPRS Vaihingen (1 \% labeled data). The RMSEs are reported in meters for all pixels (Total), for each land cover type, and for buildings averaged by different heights (Building Balanced). Building Relative denotes the mean relative error of building instances. The self-training pipelines are denoted in the format {Teacher}-{Student}, where the teacher and student networks are selected from: \texttt{reg.} (plain regression network), \texttt{cls.} (plain classification network), and \texttt{regcls.} (multi-task regression-classification network). For all pipelines except TSE, the teacher network is used for inference; in TSE, however, the exam network---defined as the EMA of the student---is used. Classification discretization schemes include UD (uniform discretization) and SID (space-increasing discretization).}
    \resizebox{\textwidth}{!}{%
    \begin{tabular}{c|c|c|ccccc|cc}
    \toprule
         Category & Configuration & Total & Impervious & Building & Low Vegetation & Tree & Car & \multicolumn{2}{c}{Building} \\
         & & & & & & & & Balanced & Relative \\ \midrule
         Self- & reg.-reg. & 3.5690 & 1.4624 & 5.4670 & 2.5674 & 3.6195 & 1.8846 & 7.7809 & 0.6051 \\
         training & regcls.-regcls. & 4.0235 & 1.6941 & 4.8137 & 5.2444 & 3.8501 & 1.7629 & 7.0092 & 0.4931 \\
         Pipeline & regcls.-cls. & 3.5718 & 1.9335 & 4.8467 & 3.2681 & 3.7583 & 2.8338 & 6.7922 & 0.4870 \\
         & regcls.-reg. & 3.5148 & 2.0206 & 4.8726 & 2.9134 & 3.7416 & 2.2245 & 6.9466 & 0.5216 \\
         & regcls.-reg. (TSE) & 3.4185 & 1.8135 & 4.8239 & 2.8901 & 3.5589 & 2.1243 & 6.6833 & 0.4737 \\ \midrule
         Classi- & UD & 3.6065 & 1.6361 & 5.0591 & 3.3448 & 3.7305 & 2.1987 & 6.9318 & 0.5172 \\
         fication & SID & 3.5476 & 1.6688 & 4.8980 & 3.4198 & 3.6494 & 2.0664 & 6.9178 & 0.5269 \\ \midrule
         Pseudo- & w/o PL loss & 3.4655 & 1.4674 & 4.9928 & 2.8346 & 3.7535 & 1.6629 & 7.1342 & 0.5106 \\ 
         label & w/o ranking & 3.5520 & 1.7677 & 4.7392 & 3.5890 & 3.6755 & 2.1361 & 6.8539 & 0.4474 \\
         Filtering & w/o dynamic thres. & 3.6958 & 1.8443 & 5.2492 & 3.2057 & 3.5340 & 6.2155 & 7.0270 & 0.5278 \\ \midrule \midrule
         \multicolumn{2}{c|}{Full Configuration} & 3.4485 & 1.5363 & 5.0094 & 2.8263 & 3.6228 & 1.9090 & 7.1249 & 0.5495 \\ \bottomrule
    \end{tabular}}
    \label{tab:vai_abl}
\end{table*}
\begin{table*}[!htbp]
    \centering
    \footnotesize
    \caption{Ablation Studies on SynRS3D (1 \% labeled data). The RMSEs are reported in meters for all pixels (Total), for each land cover type, and for buildings averaged by different heights (Building Balanced). Building Relative denotes the mean relative error of building instances. The self-training pipelines are denoted in the format {Teacher}-{Student}, where the teacher and student networks are selected from: \texttt{reg.} (plain regression network), \texttt{cls.} (plain classification network), and \texttt{regcls.} (multi-task regression-classification network). For all pipelines except TSE, the teacher network is used for inference; in TSE, however, the exam network---defined as the EMA of the student---is used. Classification discretization schemes include UD (uniform discretization) and SID (space-increasing discretization).}
    \resizebox{\textwidth}{!}{%
    \begin{tabular}{c|c|c|cccccccc|cc}
    \toprule
         Category & Configuration & Total & Bareland & Rangeland & Developed & Road & Tree & Water & Agriculture & Building & \multicolumn{2}{c}{Building} \\
         & & & & & Space & & & & Land & & Balanced & Relative \\ \midrule
         Self- & reg.-reg. & 12.1200 & 2.6334 & 2.7319 & 8.4961 & 3.5532 & 5.0057 & 1.9659 & 2.0365 & 27.5353 & 31.5760 & 0.7655 \\
         training & regcls.-regcls. & 15.5575 & 4.8723 & 3.2632 & 11.9069 & 6.1186 & 5.8625 & 4.1662 & 4.1292 & 35.0291 & 36.2192 & 0.8233 \\
         Pipeline & regcls.-cls. & 15.4924 & 6.2717 & 4.5106 & 12.6989 & 6.4734 & 6.2869 & 6.5448 & 5.4393 & 34.0093 & 35.4422 & 0.8434 \\
         & regcls.-reg. & 12.9657 & 3.2598 & 3.1848 & 8.9516 & 4.3149 & 5.1773 & 2.3070 & 2.6106 & 29.4391 & 32.6846 & 0.7659 \\
         & regcls.-reg. (TSE) & 13.6823 & 3.4587 & 2.6387 & 8.4076 & 4.5235 & 5.1789 & 2.1375 & 2.4785 & 31.4211 & 34.2528 & 0.7618 \\ \midrule
         Classi- & UD & 13.4664 & 4.3846 & 2.7773 & 8.8040 & 4.1998 & 5.7417 & 3.7274 & 2.9660 & 30.5592 & 35.2854 & 0.7899 \\
         fication & SID & 13.8618 & 4.2508 & 2.6542 & 10.0538 & 4.1653 & 5.6254 & 3.4102 & 3.4820 & 31.3642 & 35.1998 & 0.7771 \\ \midrule
         Pseudo- & w/o PL loss & 14.0690 & 3.8011 & 2.8599 & 9.3863 & 4.3254 & 5.4906 & 3.2674 & 3.4607 & 32.0461 & 35.5205 & 0.7683 \\ 
         label & w/o ranking & 13.4925 & 2.5492 & 2.3617 & 7.0471 & 3.6027 & 5.0663 & 1.6986 & 1.9790 & 31.3053 & 34.6684 & 0.7469 \\
         Filtering & w/o dynamic thres. & 13.1779 & 3.8301 & 2.7317 & 8.4527 & 3.8799 & 5.2037 & 2.0976 & 2.4370 & 30.1262 & 33.8060 & 0.7431\\ \midrule \midrule
         \multicolumn{2}{c|}{Full Configuration} & 14.1046 & 4.6899 & 2.9150 & 10.7659 & 4.5643 & 5.6551 & 3.7841 & 3.4999 & 31.7323 & 35.0314 & 0.7961 \\ \bottomrule
    \end{tabular}}
    \label{tab:syn_abl}
\end{table*}
In this section, ablation studies on the self-training pipeline, the classification process, and the pseudo-label filtering process are conducted, and the results are reported in Table \ref{tab:vai_abl}, Table \ref{tab:syn_abl}, and Table \ref{tab:gbh_abl}. 

\begin{table*}[!htbp]
    \centering
    \footnotesize
    \caption{Ablation Studies on GBH (1 \% labeled data). The RMSEs are reported in meters for all pixels (Total), for each land cover type, and for buildings averaged by different heights (Building Balanced). Building Relative denotes the mean relative error of building instances. The self-training pipelines are denoted in the format {Teacher}-{Student}, where the teacher and student networks are selected from: \texttt{reg.} (plain regression network), \texttt{cls.} (plain classification network), and \texttt{regcls.} (multi-task regression-classification network). For all pipelines except TSE, the teacher network is used for inference; in TSE, however, the exam network---defined as the EMA of the student---is used. Classification discretization schemes include UD (uniform discretization) and SID (space-increasing discretization).}
    \resizebox{\textwidth}{!}{
    \begin{tabular}{c|c|c|cc|cc}
    \toprule
         Category & Configuration & Total & Non-buidling & Building & \multicolumn{2}{c}{Building} \\
         & & & & & Balanced & Relative \\ \midrule
         Self- & reg.-reg. & 6.4801 & 4.7558 & 11.0103 & 44.8061 & 0.5024 \\
         training & regcls.-regcls. & 6.7891 & 4.6655 & 12.0655 & 46.1828 & 0.5870 \\
         Pipeline & regcls.-cls. & 6.8870 & 4.7023 & 12.2874 & 46.8149 & 0.6321 \\
         & regcls.-reg. & 6.7744 & 4.7510 & 11.8855 & 46.1175 & 0.6095 \\
         & regcls.-reg. (TSE) & 6.4563 & 4.7717 & 10.9105 & 44.1571 & 0.5946 \\ \midrule
         Classi- & UD & 6.3569 & 4.6251 & 10.8718 & 44.0157 & 0.5619 \\
         fication & SID & 6.2037 & 4.6628 & 10.3426 & 42.5714 & 0.4922 \\ \midrule
         Pseudo- & w/o PL loss & 6.3125 & 4.8207 & 10.3813 & 43.1037 & 0.4939 \\ 
         label & w/o ranking & 6.4187 & 4.6608 & 10.9934 & 44.5454 & 0.5973 \\
         Filtering & w/o dynamic thres. & 6.6426 & 4.6827 & 11.6146 & 45.5521 & 0.5843 \\ \midrule \midrule
         \multicolumn{2}{c|}{Full Configuration} & 6.2725 & 4.8843 & 10.1108 & 42.4920 & 0.5078\\ \bottomrule
    \end{tabular}}
    \label{tab:gbh_abl}
\end{table*}
\subsection{Self-training Pipeline}
The self-training pipeline represents a state-of-the-art semi-supervised learning framework, typically involving a teacher and a student network. The student is updated through gradient descent, while the teacher is maintained as the EMA of the student. Labeled data directly supervise the student, whereas pseudo-labels generated by the teacher from weakly augmented views guide the student on strongly augmented views. Consequently, a pseudo-label filtering process is essential to discard unreliable pseudo-labels and stabilize the teacher's performance during inference.

To facilitate pseudo-label filtering, we introduce a classification branch as an auxiliary task that provides a proxy for assessing the reliability of regression pseudo-labels. As reported in Tables \ref{tab:vai_abl}--\ref{tab:gbh_abl}, the \texttt{regcls.-reg.} configuration achieves consistently superior performance compared with other pipelines that incorporate classification, except for on SynRS3D, where the \texttt{reg.-reg.} configuration remains competitive.

Furthermore, our experiments reveal two notable distinctions from conventional self-training frameworks designed for classification or segmentation tasks. First, both the teacher and the student must be initialized with supervision from labeled data; without such initialization, training becomes unstable and fails to converge. Second, maintaining an additional “exam” model defined as the EMA of the student yields more reliable predictions than directly relying on the teacher during inference. We denote this configuration as the Teacher–Student–Exam (\texttt{TSE}) pipeline, which produces to systematic improvements on ISPRS Vaihingen and GBH.

The advantages of TSE arise from two main factors. On the one hand, although the classification branch supports pseudo-label filtering, it can still interfere with the regression objective and degrade performance. On the other hand, when the teacher and student architectures differ, EMA updates do not guarantee optimal knowledge transfer. In such cases, the exam model functions as a held-out “teacher” sharing the same architecture and objective as the student. By continuously accumulating the student’s improvements through EMA, the exam model stabilizes training and maximizes the benefits of self-training during inference.

On the SynRS3D dataset, with 1\% labels, the simple \texttt{reg.-reg.} pipeline already achieves competitive results. We attribute this to two factors: first, SynRS3D is a high-fidelity synthetic dataset, which facilitates pattern learning and produces reliable pseudo-labels from the teacher network. Second, the labeled split exhibits a markedly different distribution from the unlabeled, validation, and test sets. In this scenario, regression networks demonstrate stronger extrapolation capabilities than classification networks, which are less effective under distributional shifts. Notably, even though the full configuration is slightly suboptimal on SynRS3D, it still substantially outperforms the purely supervised baseline, confirming the utility of self-training in this context.
\subsection{Classification}
As discussed above, the classification branch underpins the pseudo-label filtering process, making its design critical to overall performance. A key consideration is how to discretize the continuous height values into classes. To this end, we compare several discretization strategies, as illustrated in Fig.~\ref{fig:bins}.

From an implementation perspective, Adaptive Bins (AdaBins) defines different class boundaries for each image, which complicates cross-image comparison of class confidence and therefore undermines its suitability for pseudo-label filtering. Uniform discretization (UD) and space-increasing discretization (SID), by contrast, apply fixed binning schemes but disregard the empirical distribution of height values. When the labeled dataset is small, this often results in empty classes, increasing computational overhead without improving learning.

The results demonstrate that Hierarchical Bi-Cut (HBC) provides the most consistent performance. On both ISPRS Vaihingen and GBH, HBC achieves the lowest overall RMSEs, while also yielding balanced building RMSEs that are competitive with or superior to those obtained with UD and SID (with the only exception being a marginal drop on ISPRS Vaihingen). These findings confirm the HBC serves as a principled discretization strategy, offering both a robust class definition for pseudo-label filtering and moderate yet consistent performance gains.
\subsection{Pseudo-label Filtering}
Pseudo-label filtering is critical for maintaining the reliability of pseudo-labels used to train the student network. To this end, we introduce a Plackett-Luce (PL) model that explicitly aligns class probabilities with regression errors. By ranking pseudo-labels according to their predicted class confidence, the PL model enables the systematic removal of uncertain labels and thus stabilizes the learning process.

Experimental results show that incorporating the PL model consistently improves overall RMSE and class-specific RMSEs, particularly for tall land-cover categories and balanced building metrics across all datasets. These improvements demonstrate that the PL model effectively connects classification confidence with regression accuracy, especially benefiting rare or tail pixels that are otherwise susceptible to bias. Such gains enhance robustness and promote fairer performance across diverse height ranges.

However, the PL model alone is insufficient. Without ranking-based filtering, performance degrades substantially on ISPRS Vaihingen and GBH, with SynRS3D being the only exception. This suggests that while the PL model establishes a useful correlation between probabilities and errors, it can slightly impair regression performance if applied indiscriminately. Ranking-based filtering thus plays a complementary role in effectively leveraging the PL model.

Finally, because the reliability of pseudo-labels evolves during training, we employ a dynamic thresholding strategy to regulate the filtering schedule. Early in training, when the teacher lacks sufficient knowledge, only a few pseudo-labels are admitted, preventing noise from corrupting the student’s learning. As training progresses and the teacher improves, the threshold adapts to admit more pseudo-labels, thereby accelerating convergence with increasingly reliable supervision. Empirical evidence confirms the importance of this mechanism: without dynamic thresholds, ranking filtering fails entirely on ISPRS Vaihingen and GBH, highlighting its critical role in enabling pseudo-label filtering to function as intended.

\section{Conclusions}\label{chap:conclusions}
In this paper, we presented TSE-Net, a self-training-based semi-supervised framework for monocular height estimation, representing the first work to address semi-supervised learning in pixel-wise regression tasks. To achieve this, we employ a multi-task regression-classification network as the teacher to facilitate pseudo-label filtering, along with a separate exam network sharing the same architecture as the student to accumulate improvements derived from student training. The teacher network incorporates a classification sub-network, with classes defined via a hierarchical bi-cut to reduce bias and enhance overall performance. Furthermore, a Plackett-Luce model aligns the pseudo-label filtering proxies, namely class probabilities, with regression errors, while a ranking-based filtering process with a dynamically evolving threshold ensures the reliability of pseudo-label selection.

As a prototypical study, our experiments demonstrate the potential of semi-supervised learning for regression tasks, achieving a performance gap reduction of up to 29.07\% between models trained on 100\% labeled data and only 0.1\% labeled data. These results suggest that TSE-Net can be particularly beneficial for large-scale studies towards global 3D building modeling \cite{global3d} where labeled data are scarce, enabling effective model training with minimal supervision. Future work should further investigate the impact of distributional differences between labeled and unlabeled datasets on performance and explore strategies to further enhance semi-supervised regression in diverse scenarios.
\bibliographystyle{elsarticle-num} 
\bibliography{ISPRS2025}
\end{document}